\documentclass[journal,twoside,final,letterpaper,twocolumn]{IEEEtran}

\usepackage{pifont} 
\newcommand{\cmark}{\ding{51}} 
\newcommand{\xmark}{\ding{55}} 

\ifCLASSINFOpdf
\usepackage[pdftex]{graphicx}
\else

\fi

\usepackage{epstopdf}
\usepackage{epsfig}
\usepackage{float}
\usepackage{amssymb,amsbsy,subfigure,amsmath,amsfonts,multirow,color,url,bm,cite,cases,eqparbox}
\usepackage{algorithm}
\usepackage{algpseudocode}

\usepackage[mathscr]{eucal}
\usepackage{amssymb}
\usepackage{soul}
\usepackage{xcolor}

\def\0{\mathbf{0}}
\def\1{\mathbf{1}}

\def\H{\mathbf{H}}
\def\K{\mathbf{K}}

\def\Q{\mathbf{Q}}
\def\P{\mathbf{P}}

\def\S{\mathbf{S}}

\def\V{\mathbf{V}}
\def\W{\mathbf{W}}
\def\X{\mathbf{X}}
\def\Y{\mathbf{Y}}

\def\b{\mathbf{b}}
\def\c{\mathbf{c}}
\def\d{\mathbf{d}}

\def\m{\mathbf{m}}

\def\r{\mathbf{r}}

\def\u{\mathbf{u}}
\def\v{\mathbf{v}}

\def\x{\mathbf{x}}

\def\1{\mathbf{1}_{I\times J}}

\def\Rbb{\mathbb{R}}

\usepackage{tabularx}
\usepackage{booktabs}
\usepackage{multirow}

\begin{document}
	
	\title{SUIT: Spatial-Spectral Union-Intersection Interaction Network for Hyperspectral Object Tracking}
	
	\author{Fengchao Xiong, \IEEEmembership{Member,~IEEE}, Zhenxing Wu, Sen Jia, \IEEEmembership{Senior Member,~IEEE} and Yuntao Qian, \IEEEmembership{Senior Member,~IEEE}
		\thanks { This work was supported in part by the National Natural Science Foundation of China under Grant 62371237 and in part by
		the Fundamental Research Funds for the Central Universities under Grant
		30923010213.}
		\thanks{F. Xiong, and Z. Wu are with the School of Computer Science and Engineering, Nanjing University of Science and Technology, Nanjing 210094, P.R. China. Corresponding author: F. Xiong(fcxiong@njust.edu.cn). }
		\thanks{S. Jia is with  College of Computer Science and Software Engineering, Shenzhen University, Shenzhen 518060, P.R. China.}
		\thanks{Y. Qian  is with the College of Computer Science, Zhejiang University, Hangzhou 310027, China.}
	}
	\maketitle

	\begin{abstract}Hyperspectral videos (HSVs), with their inherent spatial-spectral-temporal structure, offer distinct advantages in challenging tracking scenarios such as cluttered backgrounds and small objects. However, existing methods primarily focus on spatial interactions between the template and search regions, often overlooking spectral interactions, leading to suboptimal performance. To address this issue, this paper investigates spectral interactions from both the architectural and training perspectives. At the architectural level, we first establish band-wise long-range spatial relationships between the template and search regions using Transformers. We then model spectral interactions using the inclusion-exclusion principle from set theory, treating them as the union of spatial interactions across all bands. This enables the effective integration of both shared and band-specific spatial cues.   At the training level, we introduce a spectral loss  to enforce material distribution alignment between the template and predicted regions, enhancing robustness to shape deformation and appearance variations.   Extensive experiments demonstrate that our tracker achieves state-of-the-art tracking performance. The source code, trained models  and  results  will be publicly available via https://github.com/bearshng/suit to support reproducibility.

		
	\end{abstract}
	\section{Introduction}
	
	\begin{figure}[!ht]
	\begin{center}
		\includegraphics[width=\columnwidth]{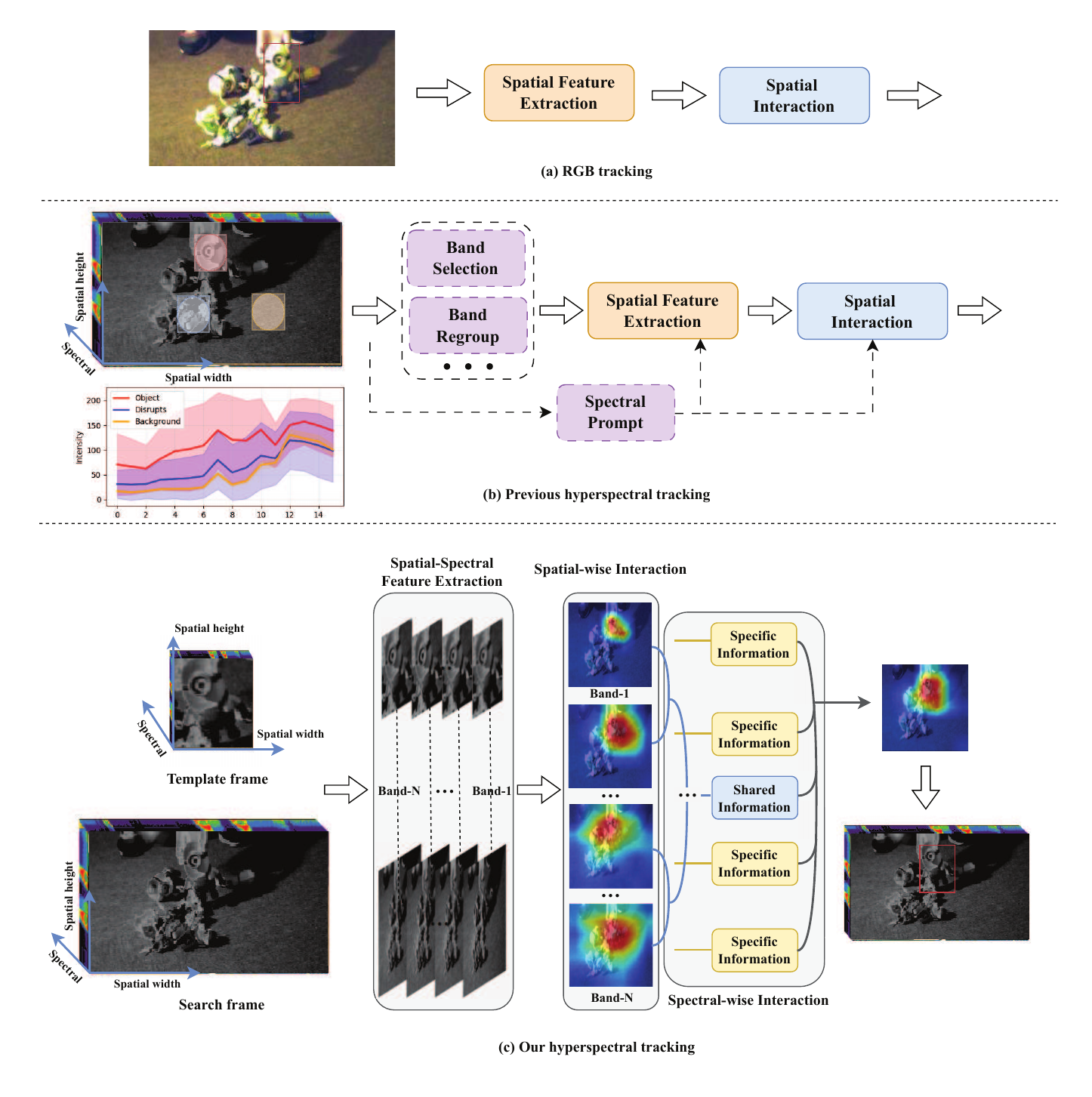}
	\end{center}
	\caption{(a) RGB tracking relies solely on spatial feature extraction and spatial-wise interaction. (b) Previous hyperspectral tracking methods adapt RGB tracking pipelines by applying band selection or grouping, followed by spatial-wise interaction, often ignoring spectral relationships. (c) Our proposed SUIT tracker fully leverages the spatial-spectral-temporal structure inherent to hyperspectral videos by explicitly modeling spatial-spectral interactions, enabling more effective correlation modeling for robust tracking.}\label{Tracking_Method}
\end{figure}

	Visual object tracking is a fundamental task in computer vision, with wide-ranging applications such as autonomous driving and smart city development~\cite{javed2022visual, Zheng2024, Guo2021, Li2023, Tang2024, Luo2024, Shao2021HRSiam}. While RGB-based trackers have achieved significant progress, they often struggle in complex scenarios such as background clutter, occlusion, and similar object appearances. This limitation arises from the inherent ambiguity in RGB features, which lack material-level discrimination. Hyperspectral videos (HSVs), characterized by their rich spatial-spectral-temporal structure, offer a compelling alternative. The high spectral resolution enables robust material identification, making HSVs particularly advantageous in scenarios involving occlusion, deformation, or significant background interference~\cite{Xiong20MHT, Liu22SiamHYPER}. Consequently, hyperspectral object tracking has attracted increasing interest, evolving from traditional handcrafted feature methods~\cite{Xiong20MHT, Hou22TSCEW, zhang2023fast} to deep learning-based approaches~\cite{Li20BAE-Net, Liu22SiamHYPER, Wang22SSATFN, Liu22H3,Wang2025b}.

	Despite these advances, most existing hyperspectral trackers remain limited by their architectural similarity to RGB trackers. As illustrated in Fig.~\ref{Tracking_Method} (a),  in RGB tracking, spatial features are extracted from both the template and search regions, followed by spatial-wise interaction to establish similarity for tracking. Inspired by this pipeline, as given in Fig.~\ref{Tracking_Method} (b), prior hyperspectral tracking methods typically perform band selection or band regrouping to convert hyperspectral videos into multiple false-color groups. These groups are then processed using adapted RGB tracking frameworks. As a result, they focus heavily on spatial interactions between the template and search regions—using modules like Transformers~\cite{Gao2023a} or correlation networks~\cite{Zhao2022},while largely ignoring the spectral-wise interactions that are essential for leveraging the full potential of hyperspectral data.  Neglecting the spectral dimension leaves valuable material cues unexploited and limits tracking robustness.

	Another under explored but critical challenge in HSV tracking is robustness under spatial ambiguity. When the target undergoes deformation, blurring, or rapid motion, spatial cues can become unreliable. However, the underlying material distribution remains relatively stable, offering an alternative cue for object consistency. Unfortunately, few existing methods explicitly exploit this property for tracking enhancement.

We argue that the key to unlocking the potential of HSVs lies in explicitly modeling spatial-spectral interactions. To this end, we introduce the spatial-spectral union-intersection interaction network, abbreviated as SUIT, for hyperspectral object tracking. SUIT is designed to explicitly model the spatial-spectral interaction structure from both architectural and training perspectives. From the architectural aspect, as in Fig.~\ref{Tracking_Method} (c), SUIT treats HSVs as stacks of grayscale images across bands and first establishes band-wise spatial interactions between the template and search regions using multi-head self- and cross-attention.
While all bands capture the same object, they reflect different physical properties across wavelengths. This implies that there are both shared and band-specific interactions between the template and search patches. Therefore, drawing from set theory, the network then integrates these interactions using the inclusion-exclusion principle, where band-shared and band-specific interactions are disentangled and fused to construct a comprehensive spatial-spectral interaction.  From the training aspect, we introduce a novel spectral loss that enforces material distribution consistency between the template and predicted regions. This encourages the network to focus on material cues that remain stable under deformation, blur, or occlusion, thereby improving robustness in complex scenarios. Extensive experiments show that our SUIT tracker achieves robust hyperspectral tracking.

The contributions of this paper can be summarized as follows:	
	\begin{itemize}
		\item We study hyperspectral tracking from a novel paradigm by focusing on spatial-spectral interaction modeling in hyperspectral videos, rather than only enhancing spatial matching as in previous RGB-adapted approaches.
		\item We propose a novel spatial-spectral interaction network and introduce an effective fusion approach for spectral-wise interaction grounded in the inclusion-exclusion principle from set theory. This method treats the integration of   spectral-wise interaction as the union of spatial interactions across all bands to reliably extract both shared and specific interactions among bands for robust fusion.
		\item We introduce a novel spectral loss to enforce the material distribution consistency between the search and template regions, improving the tracking robustness to  unexpected spatial changes.
	\end{itemize}
	
	\begin{figure*}[t]
		\begin{center}
			\includegraphics[width=\textwidth]{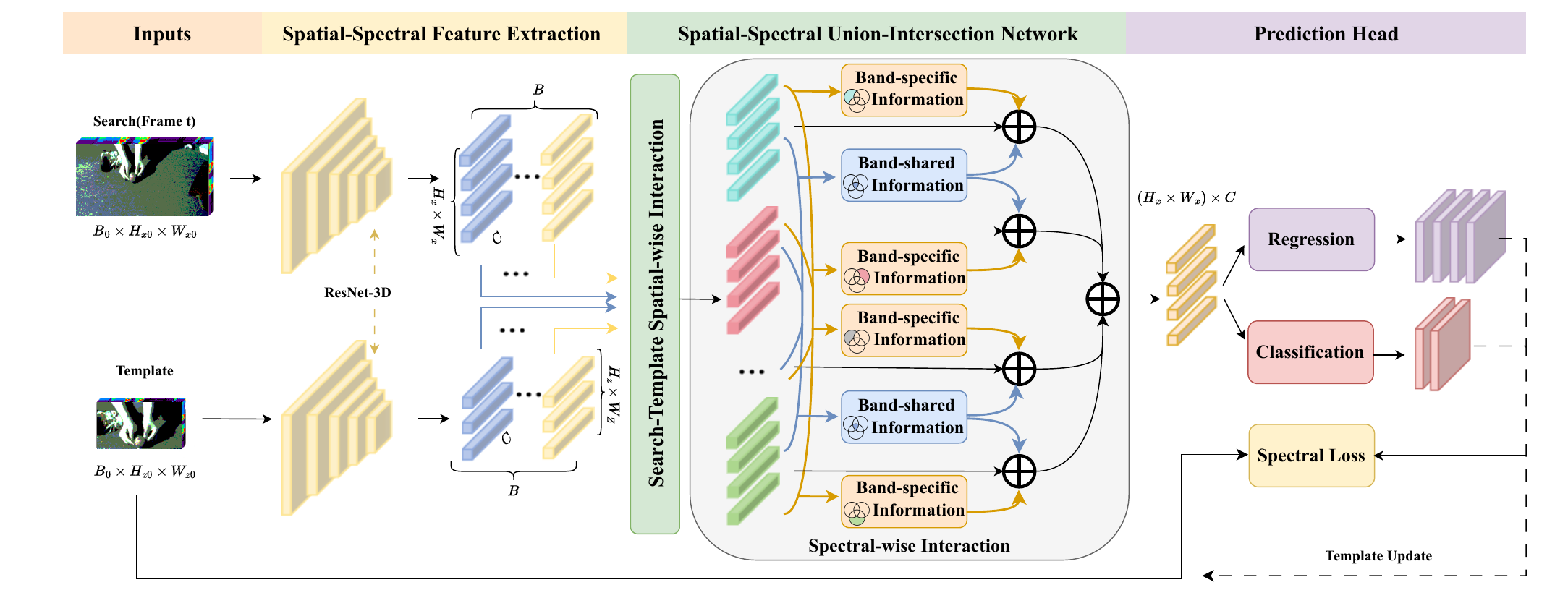}
		\end{center}
		\caption{Architecture of our spatial-spectral union-intersection
		interaction tracker. The  tracker contains three fundamental components: spatial-spectral feature extraction backbone, spatial-spectral union-intersection  network, and prediction head.}
		\label{Architecture}
	\end{figure*}

	The remainder of this paper is organized as follows: Section~\ref{sec:rec} introduces the related works. Section~\ref{sec:method} gives the details of the proposed tracker. Section~\ref{sec:exp} compares the proposed SUIT with alternative trackers. Section~\ref{sec:dis} presents a discussion of this paper. The conclusions are drawn in Section~\ref{sec:con}.

	\section{Related Work} \label{sec:rec}
	
	This section introduces the recent advances of hyperspectral object tracking and the fusion techniques in object tracking.
	\subsubsection{Hyperspectral Object Tracking}
Hyperspectral data captures detailed physical spectral information across a wide range of wavelengths and possesses strong capabilities for material identification~\cite{Li2023a, Hong2024, Li2025}. Traditionally, the acquisition of hyperspectral data has been limited by low imaging speeds, making it difficult to obtain sufficient hyperspectral video data. However, with recent advancements in sensor technology and imaging mechanisms, it has become increasingly feasible to capture hyperspectral videos using commercially available cameras. Nevertheless, such data typically requires professional postprocessing steps, including spectral correction, to ensure accuracy and reliability. For this reason, hyperspectral object tracking suffers from ``data hungry" problem.  Therefore, early hyperspectral trackers~\cite{Xiong20MHT,Hou22TSCEW,Tang2022} relied on handcrafted features to describe the spatial-spectral property of the objects.  The captured hyperspectral data forms a continuous curve capable of identifying different materials. Uzkent~\emph{et al.}~\cite{Uzkent2016} computed local hyperspectral histograms to improve tracking performance under varying lighting conditions. While spectral features alone can be effective, incorporating local spatial structure information is essential for addressing complex tracking scenarios. To achieve this, Hou~\emph{et al.}~\cite{Hou22TSCEW} employed a band-by-band histogram of oriented gradients (HOG) feature extractor, combining it with intensity HOG for enhanced spatial feature representation. Xiong~\emph{et al.}~\cite{Xiong20MHT} calculated the spectral-spatial histogram of multidimensional gradients (SSHMG) to encode 3D local spectral-spatial structures in hyperspectral videos and  further integrated these features with fractional abundances, thereby enhancing tracking robustness.

	Since the introduction of large-scale hyperspectral tracking dataset by   Xiong~\emph{et al.}~\cite{Xiong20MHT}, hyperspectral tracking has shifted to deep learning (DL)-based ones~\cite{Liu2024,Yao2025}. A DL-tracker generally comprises three components: the backbone network for appearance representation, an interaction network to establish relationships between the template and search regions, and a prediction head for classification and regression. Given no availability of backbone tailored for hyperspectral images, many existing works leverage  RGB-based backbone networks pre-trained on large-scale color datasets for appearance representation. Based on the concept of band selection, approaches like BAE-Net~\cite{Li20BAE-Net}, SEE-Net~\cite{Li23SEE-Net},SSTtrack~\cite{Chen2025b}, and SiamBAG~\cite{li2023siambag} divided hyperspectral frames into multiple groups of false-color frames accoridng to their band importance for representation which were then integrated at the decision level for tracking. Li~\emph{et al.}~\cite{Li23SEE-Net} advanced this idea by establishing a spectral self-expression model to learn band correlations and generated weighted false color frames for robust object tracking. Chen~\emph{et al.}~\cite{Chen2024} further improved the tracking performance by introducing the  inter and intra band feature interactions for fusing the extracted features of multiple groups.   Additionally, Transformers have been employed in~\cite{Chen2023, Chen2025} to achieve long-range feature extraction. Moreover, some approaches convert HSIs into pseudo-RGB videos and fuse these with RGB videos for multimodal RGB-Hyperspectral tracking~\cite{Zhao2022,Liu2021,Sun2023}
	To address the challenge of scale variation,  Zhao~\emph{et. al}~\cite{Zhao2025} introduced a scale aware update module and a segmented template update strategy. Prompt tuning strategies were introduced in~\cite{Chen2025a} to adapt RGB-trackers for hyperspectral tracking. 
	
	

	As for  interaction, almost all the methods are directly adapted from the RGB tracking. The correlation-based networks that build the spatial-wise similarity between the search region and template region is typically used~\cite{Li2024}. Driven by the success of TransT~\cite{chen2021transt},~\cite{Gao2023,Wu2024,Wang2025,Wang2025a,wang2023spectral} adopted Transformer-based fusion module to integrate the features of template and search patches for tracking.  Concretely, Wu~\emph{et. al}~\cite{Wu2024} incorporated a domain adaptive-aware module preceding the Transformer-based framework, subsequently conducting spatial feature interaction to integrate the features of template and search patches for tracking. Gao~\emph{et. al}~\cite{Gao2023} developed a bidirectional multiple deep feature fusion module and a cross-band group attention module to enhance interaction information among generated false-color frames. Jiang~\emph{et. al}~\cite{jiang2024siamcat} introduced a channel-adaptive dual Siamese network, leveraging band-domain adaptation followed by spatial feature interaction.  To address the overfitting issue commonly observed in Transformer-based trackers on small-scale hyperspectral video datasets, Wang~\emph{et. al}~\cite{Wang2025}  introduced an adaptive content and position embedding module   to   dynamically balance the importance between positional information and content-based features. 
	
	However, all these works only perform interaction in the spatial domain but fails to consider that in the spectral domain.  The main reason is that all these methods make exhaustive efforts on the spatial-domain feature extraction and produces  features typically three-dimensional, encompassing channel, height, and width, but lacking the band dimension. In this paper, we perform joint spatial-spectral feature extraction to produce four-dimensional feature maps, encompassing channels, bands, height, and width. We then apply spatial-spectral interaction at the architecture level and spectral loss at the training level  between regions to enhance tracking.
	
	\subsubsection{Fusion in Object Tracking}
	Fusion techniques for object tracking have become essential in enhancing tracking performance and robustness under challenging conditions by leveraging complementary information from multiple sources~\cite{wang2024MultiModalFusion}. There are three main types of fusion algorithms based on deep learning: pixel-level fusion, decision-level fusion, and feature-level fusion.  Zhang~\emph{et al.}\cite{Zhang2019} adopted a pixel-level fusion approach using a Siamese network to combine RGB and infrared modality information. Unlike pixel-level fusion, decision-level fusion combines the tracking results of different modalities at the output stage. Several works\cite{lan2019DecisionTracker, Li23SEE-Net} employ multiple trackers to track multi-modal data, determining the fusion results by weighting or selecting the maximum confidence score.
	
	Feature-level fusion has gained popularity due to its ability to effectively harness and combine complementary information of different sources.  Ma~\emph{et. al}~\cite{ma2015hierarchical}, Li~\emph{et. al}~\cite{li2019siamrpn++}, and Zhang~\emph{et. al}~\cite{zhang2020dsiammft}  have achieved improved tracking performance by adaptively fusing convolutional features from multimodal images. Many studies have explored  multi-scale techniques~\cite{li2023multiScale} and attention mechanisms~\cite{guo2021graph, yu2020deformable} to enrich the semantic content of target features. Furthermore, the advent of Transformers~\cite{vaswani2017attention} has enabled trackers~\cite{chen2021transt, ye2022joint, cao2024BiDirectional} to exploit complex and hierarchical feature interactions, leading to significant performance improvements. Some Transformer-based trackers~\cite{cao2021hift, chen2022efficient, yan2021learning, lin2022swintrack} utilize CNN-based backbones for feature extraction while leveraging Transformers to model relationships between the search and template branches.
	
	In this paper, we leverage Transformers to integrate the template and search regions across both spatial and spectral domains. To fully harness the spectral information, we treat spectral-wise interaction as the union of spatial interaction  across all bands. This process is guided by the inclusion-exclusion principle from set theory, ensuring a more comprehensive and effective integration with higher guarantees.

	\section{Proposed Method}\label{sec:method}	
	This section introduces the details of our method, encompassing the overall network, the introduced spatial-spectral union-intersection network and  spectral  loss function.
	
	\subsection{The Architecture}
	
	As illustrated in Fig.~\ref{Architecture}, our tracker employs a Siamese architecture with two branches: the template branch and the search branch. Both branches share the same structure and weights, processing the template patch $z \in \mathbb{R}^{B_0\times H_{z_0}\times W_{z_0}}$ and the search patch  $x \in \mathbb{R}^{B_0\times H_{x_0}\times W_{x_0}}$, respectively. Each branch consists of a feature extraction network, a spatial-spectral union-intersection network, and a prediction head. The feature extraction network  is responsible for extracting local spatial-spectral features to describe the object. Specifically, the network extends the ResNet-50 architecture~\cite{he2016resnet}, originally trained on color images, by expanding the 2D convolutions to asymmetric 3D convolutions.  The 3D convolution  decouples the parameters from the number of bands, enabling the network to accept HSVs with an arbitrary number of bands. In detail, we replace the conventional $3\times 3\times 3$    convolution with an asymmetric convolutional scheme, where the $1\times 3\times 3$ convolution primarily extracts spatial information and the $3\times 1\times 1$ convolution extracts the spectral information.    Additionally, pooling is performed in both the spatial and spectral domains to enlarge the spatial-spectral receptive field. The last layer of the original architecture is removed, utilizing the fourth stage as the output.   With the  backbone network,  the search region and the template are represented as   $F_x \in \mathbb{R}^{C \times B \times H_x \times W_x}$ and
	$F_z \in \mathbb{R}^{C \times B \times H_z \times W_z}$, corresponding to the channel, band, height and width, respectively. Here, $H_z, W_z = \frac{H_{z0}}{8}, \frac{W_{z0}}{8}$, $H_x, W_x = \frac{H_{x0}}{8}, \frac{W_{x0}}{8}$, $B = \frac{B_0}{4}$ and $C=1024$.  We then build the relationship between $F_z$ and $F_x$ with spatial-spectral union-intersection network.  The output is subsequently processed by the head network, which generates a binary classification map and a regression map for tracking. The prediction head comprises a classification branch and a regression branch, each consisting of a three-layer perceptron with  a ReLU activation function.
	\subsection{Spatial-Spectral Union-Intersection  Network}\label{subs:SUIT}

Before delving into the details of our spatial-spectral interaction, we first define the fundamental component: multihead attention.  Given the queries $\Q\in \mathbb{R}^{N\times d}$, keys $\K \in \mathbb{R}^{N\times d}$ and values $\V \in \mathbb{R}^{N\times d}$, the multihead attention  projects $\Q$, $\K$ and $\V$ for $h$ times with different parameters $\W^Q_h \in \mathbb{R}^{d\times d_k}, \W^K_h \in \mathbb{R}^{d\times d_k}, \W^V_h \in \mathbb{R}^{d\times d_v}$  and computes the attention as follows:
\begin{equation}
	\begin{split}
		&\text{MultiHead}(\Q, \K, \V) = \text{Concat}(\H_1, \cdots, \H_h)\W^O,\\\
		&\H_i = \text{Attention}(\Q\W_h^Q, \K\W_h^K, \V\W_h^V), \label{eq:mh}
	\end{split}
\end{equation}
where \text{Attention} is the scaled dot-product attention mechanism, and  $\W^O \in \mathbb{R}^{hd_v\times d}$ represents the learnable projection matrix. Based on the multihead attention, we define the following two basic blocks.

\noindent \textbf{Spatial Context Enhancer (SCE) Block:} Given feature map $\X$, the SCE block explores the non-local context  with multihead self-attention to augment the representation capabilities and is defined as
\begin{equation}
	\X_{\text{SCE}} = \X + \text{LN}(\text{MultiHead}(\X + \P_x, \X + \P_x, \X)),
\end{equation}
where $\P_{x}$ is the positional encoding using sine function that encodes spatial positions and $\text{LN}$ is the layer normalization.

\noindent \textbf{Spatial Context Fusion (SCF) Block:}  Given the
feature maps $\X$ and $\Y$, the SCF block extracts the shared information  with multihead cross-attention  for  fusion, defined by
\begin{equation}
	\begin{split}
		&\widetilde{\X}= \X + \text{FFN}(\text{MultiHead}(\X + \P_x, \Y + \P_y, \Y)),\\
		&\widetilde{\Y}= \Y + \text{FFN}(\text{MultiHead}(\Y + \P_y, \X + \P_x, \X)),\\
		&\X_{\text{SCF}}= \widetilde{\X} + \text{FFN}(\text{MultiHead}(\widetilde{\X} + \P_x, \widetilde{\Y} + \P_y, \widetilde{\Y})),
	\end{split}
\end{equation}
where  $\P_{y}$ is also  the positional encoding. $\text{FFN}$ is  a fully connected feed-forward network that consists of two linear transformation with a ReLU in between.

Now, we build the spatial-spectral interaction between the template and search regions with the introduced SCE and SCF blocks. Throughout the paper, 8 heads are adopted when using the SCE and SCF blocks.

\noindent \textbf{Spatial-wise Interaction:} Before  building the spatial interaction  between the template and search region, we first enhance the contexts within patches with the SCE blocks.  Specifically, we split $\{F_{x}, F_{z} \}$ along bands to obtain $\{F_{x_b} \in \Rbb^{C\times H_x\times W_x}, F_{z_b}\in \Rbb^{C\times H_z\times W_z}\}$ and encourage  the network to focus on the useful semantic context as follows:
\begin{equation}
	\begin{split}
		F_{x_b}&=\text{SCE}(F_{x_b})\\
		F_{z_b}&=\text{SCE}(F_{z_b})
	\end{split}
\end{equation}
Here, $\Q$ and $\K$ in the SCE module are obtained from the input features with added positional embeddings, while $\V$ is directly derived from the input. 

We then establish the spatial relationship  between the template patch and the search patch on a band-by-band basis as follows:
\begin{equation}
	f_b=\text{SCF}(F_{x_b}, F_{z_b})	
\end{equation}

\noindent \textbf{Spectral-wise Interaction:}
The generation of $f_b$ involves building the spatial relationship  between the template patch  $F_x$ and search patch $F_z$ at the $b$-th band. Since each band describes the same object but in a different wavelength, there must be some common spatial correlation information as well as distinct differences between them.  The aim of spectral-wise interaction is to simultaneously integrate such shared  and distinct spatial correlation information  for subsequent tracking.

We perform spatial-wise interactions based on the principles of set theory, where each  $f_b$ is regarded as an information set, and the spectral-wise integration is treated as the union of  $f=\{f_1, \cdots, f_b\}$.  In set theory, the inclusion-exclusion principle is a counting technique that generalizes the method for determining the number of elements in overlapping sets. The process involves first summing the individual sets, then subtracting pairwise intersections, adding back triple-wise intersections, subtracting quadruple-wise intersections, and so on. From an information processing perspective, this method is particularly effective at extracting shared information through intersection computations. By progressively removing shared information, specific information can be isolated. The joint utilization of both shared and specific information facilitates effective information fusion with greater reliability.

Therefore, based on the inclusion-exclusion principle,  the  joint utilization of $f=\{f_1, \cdots,f_b\}$ can be described as
\begin{equation}
	\begin{split}
		I\left(\bigcup_{b=1}^{B} f_b\right) &= \sum_{b=1}^{B} I(f_b) \ominus \left( \sum_{i \neq j}^{N} I\left(f_{i} \cap f_{j}\right) \right) \oplus I\left(\bigcap_{b=1}^{B} f_b\right)\\
		& =\underbrace{\bigcup_{l \neq m}^{N}\left(f_{l} \cup f_{m}\right) \ominus \bigcup_{i \neq j}^{N}\left(f_{i} \cap f_{j}\right)}_{f_{\text{specific}}} \oplus \underbrace{I\left(\bigcap_{b=1}^{B} f_b\right)}_{f_{\text{shared}}} \label{eq:set}
	\end{split}
\end{equation} where $I(\cdot)$ represents the set, $\bigcup$ represents the union operation,  $\bigcap$ represents the intersection operation, $\ominus$ computes the differences between sets, $\oplus$ merges two sets.  Here, we only consider  the pairwise intersections and $B$-tuple-wise intersection  but ignore others such as triple-wise intersections, quadruple-wise intersections and more, for simplicity.

The union operation, which aims to aggregate all information from two sets, is implemented via feature addition under the assumption of independence between the sets. To achieve this, we first isolate the shared (joint) information from each individual set, resulting in two independent sets that can then be added to fulfill the union operation.
In contrast, the intersection operation, which extracts the common information between two sets, corresponds to the function of the SCF block, capturing shared spatial correlations across spectral bands. $\ominus$ and $\oplus$ can be implemented by their definition with the subtraction and addition. 

\noindent \textbf{Band-specific  Spatial Correlation  Calculation.}
For an individual band, the band-specific component  $f_b^{\text{specific}}$ can be computed by isolating the shared  spatial correlation information  between bands from the individual  spatial correlation within bands, i.e., $f_b$. This is mathematically represented as
\begin{equation}
	f_b^{\text{specific}} = I(f_b) \ominus I\left(\bigcup_{b \neq j}^{N} f_{b} \cap f_{j}\right) \label{eq:diff}
\end{equation}
At the network level, Eq.~(\ref{eq:diff}) can be implemented as:
\begin{equation}
	f_b^{\text{specific}} = \text{SCE}(f_{b}) - \left(\text{SCE}\left(\sum_{b \neq j} \text{SCF}(f_{b}, f_{j})\right)\right)
\end{equation}
Here,   the spatial context is augmented by the SCE block  before subtraction.

\noindent \textbf{Band-shared Spatial Correlation Calculation.}   The commonality among multiple bands $f_{\text{shared}}$ is incrementally extracted through establishing the pairwise interactions between features at each band. Mathematically, the process can be represented as follows:
\begin{equation}
	\begin{split}
		f_{\text{shared}} &= \bigcap_{b=1}^{B} f_b = \bigcap_{b \neq j}^{N} \text{SCF}\left(f_b, f_j\right)  \\
		&= \text{SCF}(\text{SCF}(\cdots \text{SCF}(f_{1}, f_{2}) \cdots), f_{B})
	\end{split}	
\end{equation}
To reduce the number of parameters, all the SCF blocks share the same weights. 	

Based on the band-specific and band-shared  spatial correlation,  the aggregated  spatial correlation is obtained by following  Eq.~(\ref{eq:set}) with a residual connection:
\begin{equation}
	f_{\text{fused}} = \sum \limits_{b=1}^{B}\left(f_{\text{shared}} + f^{\text{specific}}_b + f_{b}\right)
\end{equation}
spatial correlation is obtained by establishing pairwise interactions among all bands, making it globally shared and more robust to potential disruptions at a specific band.  Therefore, we  include  $f_{\text{shared}} $ for multiple times in the interaction process. In the end, $f_{\text{fused}}$ is further enhanced by the SCE block to capture the global  spatial correlation  for subsequent tracking. Algorithm 1 summarizes  the inclusion-exclusion principle based spectral interaction.
\begin{algorithm}
    {
    \caption{Inclusion-Exclusion Principle based Spectral Interaction.}
    \begin{algorithmic}[1]
		\Require Band-wise Interaction Information $f = \{f_1, \dots, f_B\}$
		\Ensure Fused Spatial-spectral Interaction $f_{\text{fused}}$
		
		\State \textbf{Band-Shared Interaction:}
		\State $f_{\text{shared}}=f_1$
		\For{$b = 2$ to $B$}
		\State $f_{\text{shared}} \gets \text{SCF}(f_{\text{shared}}, f_b)$ \Comment{Global intersection}
		\EndFor
		\State \textbf{Band-Specific Interaction:}
		\For{$b = 1$ to $B$}
		\State $f_b^{\text{shared}} \gets \sum_{j \neq b} \text{SCF}(f_b, f_j) $ \Comment{Pairwise intersection}
		\State $f_b^{\text{specific}} \gets \text{SCE}(f_b) - \text{SCE}\left(f_b^{\text{shared}}\right)$
		\EndFor
		\State \textbf{Fusion:}
		\State $f_{\text{fused}} \gets \sum_{b=1}^B \left(f_{\text{shared}} + f_b^{\text{specific}} + f_b\right)$
		\State $f_{\text{fused}} \gets \text{SCE}(f_{\text{fused}})$
		
		\State \Return $f_{\text{fused}}$
	\end{algorithmic}

    } 
\end{algorithm}

From the design of the spatial-spectral interaction, our tracker is able to handle HSVs with an arbitrary number of bands. On one hand, the parameter decoupled 3D convolution introduces no parameters in relation to the band. On the other hand, the all the parameters in band-specific spatial relationship  information calculation  and  band-shared  spatial correlation  calculation are regardless of the number of bands.
	
	\subsection{Loss Function}\label{subs:loss_function}
	\noindent \textbf{Spectral Loss:} It is not uncommon that the object undergoes  unexpected changes in the spatial structure.  Ideally, the predicted region and the template should share a similar material distribution, regardless  of the shape. To achieve this, we introduce a spectral loss to measure the material distribution of the template and the predicted bounding box.  The template and  predicted bounding boxes usually have different sizes. One approach is to calculate the loss between average spectra, but  cannot fully represent the material properties.
		\begin{figure}[t]
		\begin{center}
			\includegraphics[width=\columnwidth]{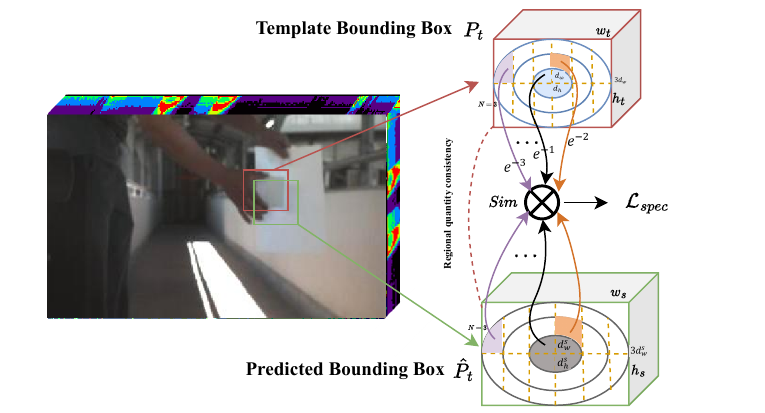}
		\end{center}
		\caption{Spectral loss function. Here we show the spectral region division and spectral loss calculation when $N=3$.}
		\label{loss}
	\end{figure}

	We instead calculate the loss at the local  level. Let $(x_t, y_t, w_t, h_t)$ represent the axis coordinates of the bounding box center, and the width and height of the template bounding box. Similarly, the predicted region is represented by $(x_s, y_s, w_s, h_s)$. With $(x_t, y_t)$ as the center, we define a series of ellipses within the template region as follows:
	\begin{equation}
		\frac{(x-x_t)^2}{(nd_w)^2} + \frac{(y-y_t)^2}{(nd_h)^2}=1, n=1,\cdots, N
	\end{equation}
	where $(d_w, d_h)$ are the preset the minimum axes length and  $N=\left\lfloor\min \left(\frac{w_{t}}{d_{w}}, \frac{h_{t}}{d_{h}}\right)\right\rfloor$ determines the number of ellipses. Let $d$ denote the  major axis length of the innermost ellipse, i.e, $d=max(d_w, d_h)$. A smaller $d$ results in overly narrow regions, where the extracted spectral curves tend to be noisy and unstable. On the other hand, a larger $d$  increases the likelihood of including mixed-material areas, which dilutes the discriminative power of the spectral features. In our experiments, we empirically set $d=5$ if $max(d_w, d_h)>5$  to maintain its locality. With $d$, the other axes can be calculated according to the aspect ratio of the template bounding box.
	Similarly, with $(x_s, y_s)$ as the center, we define ellipses within the corresponding predicted region as follows:
	\begin{equation}
		\frac{(x-x_s)^2}{(nd_w^{s})^2} + \frac{(y-y_s)^2}{(nd_h^{s})^2}=1, n=1,\cdots, N
	\end{equation}
	Here we set  $d_w^s = \frac{w_s}{N}$ and  $d_h^s=\frac{h_s}{N}$ to  ensure that each local region in the template corresponds spatially to the predicted region.
	
	Based on the defined ellipses, we first divide the template patch and predicted patch into a number of  elliptical rings, each represented as:
	\begin{equation}
		\begin{split}
			R_t &= \left\{ (x,y) | (n-1)^2< \frac{(x-x_t)^2}{(d_w)^2} + \frac{(y-y_t)^2}{(d_h)^2}\leq n^2 \right\}\\
			R_s &= \left\{ (x,y) | (n-1)^2< \frac{(x-x_s)^2}{(d_w^s)^2} + \frac{(y-y_s)^2}{(d_h^s)^2}\leq n^2 \right\}	
		\end{split}
	\end{equation}
	As show in Fig.~\ref{loss}, to further differentiate the spectra at various positions within the target, the elliptical ring is subdivided into smaller regions, we calculate the average spectral reflectance within each region to represent the material distribution. For example, for the upper half of each elliptical ring, this process is expressed by
	\begin{equation}
		\begin{split}
			\textbf{m}_t = \frac{1}{M_t}\sum_{x,y}\P_t(x,y,:)  \qquad{kd_w \leq  x \leq   (k+1)d_w}, y>0 \\
			\textbf{m}_s = \frac{1}{M_s}\sum_{x,y}\widehat{\P}_s(x,y,:)  \qquad{kd_w^s \leq x \leq (k+1)d_w^s, y>0}
		\end{split}
	\end{equation}
	where $\P_t$ and $\widehat{\P}_s$ represent the template region and predict region.  $M_t$ and  $M_s$ are the total number of pixels within the regions. By varying $k=-N\sim N-1$,  the elliptical ring is divided into distinct smaller regions. We then minimize the distance between $\m_t$ and $\m_s$:
	\begin{equation}
		\mathcal{L}_{\text{spec}} = 1-\text{exp}\left(-n\arccos\left(\frac{\textbf{m}_t^T \textbf{m}_s}{|\textbf{m}_t| \textbf{m}_s|}\right)\right)
	\end{equation}
	where $n=1\cdots N$ indexes the ellipse. Such a loss function has two merits. On one hand, calculating the loss within these smaller regions gives the tracker a certain degree of robustness against deformation challenges. On the other hand, the closer the region is to the center, the higher the loss is, which can suppress the less precise region boundaries and further improve the robustness of the tracker. By summing the loss of all the divided regions, the global spectral loss  $\mathcal{L}_{spec}$ for all the divided  regions can be obtained.
	
	Combining  cross entropy loss $\mathcal{L}_{\text{cls}}$  for classification, and IoU loss $\mathcal{L}_{\text{reg}}$ for bounding box regression, the total loss function of our SUIT  is defined by
	\begin{equation}
		\mathcal{L}  = \mathcal{L}_{\text{cls}} + \lambda_{\text{reg}} \mathcal{L}_{\text{reg}}  + \lambda_{spec} \mathcal{L}_{\text{spec}}
	\end{equation}
	where $\lambda_{\text{reg}}$ and $\lambda_{\text{spec}}$ balance the different loss functions. Here, the classification loss $\mathcal{L}_{\text{cls}}$ evaluates whether a candidate region contains the target object, formulated as
\begin{equation}
\mathcal{L}_{\text{cls}} = -\left[ y \log(p) + (1 - y) \log(1 - p) \right],
\end{equation}
where $y \in \{0,1\}$ is the ground-truth label (1 for object, 0 for background), and $p$ is the predicted probability of the target being present in the region. The IoU  penalizes misaligned bounding boxes between the predicted bounding box $\mathcal{B}_p$ and ground-truth box $\mathcal{B}_g$, computed as:
\begin{equation}
\text{IoU} = \frac{|\mathcal{B}_p \cap \mathcal{B}_g|}{|\mathcal{B}_p \cup \mathcal{B}_g|}.
\end{equation}
Then, the IoU loss  encourages high overlap between the predicted and ground-truth boxes and is  defined as:
\begin{equation}
\mathcal{L}_{\text{IoU}} = 1 - \text{IoU}.
\end{equation}

	\section{Experiment} \label{sec:exp}
	In this section, we first compare our SUIT method with state-of-the-art trackers to demonstrate its superior tracking performance.  Additionally, we conduct an ablation study to assess the effectiveness and impact of the different components on overall performance.
	\subsection{Experimental Setup}
	\noindent \textbf{Network Implementation:}
	We implemented our tracker using PyTorch and trained it on two NVIDIA 3090 GPUs. The training was conducted with the AdamW optimizer. We set the initial learning rate for the backbone network to $10^{-5}$, and $10^{-4}$ for the other parameters, with a weight decay of $10^{-4}$. The network was trained with a batch size of 16 for 200 epochs, with 1000 iterations per epoch. The learning rate was reduced by a factor of 10 after 50 epochs. The regularization parameters $\lambda_{\text{reg}}$ and $\lambda_{\text{spec}}$ were set to 2 and 1, respectively. Training a 3D ResNet can be challenging, so we focused on training the $3\times 1\times 1$ convolutional layer responsible for the spectral dimension while freezing the other convolutional layers that handle the spatial dimension. This tailored strategy effectively guides the network to learn spectral domain representations while preserving the robust spatial domain representations of the original ResNet-50.

We maintained a consistent setup across all the datasets. The data augmentation strategies included random horizontal flipping (with a probability of 0.5), random rotation (within ±15°), random cropping (while preserving the target area), and normalization.   The model was initially trained on the HOT2023 dataset, selected as the base due to its larger scale and multimodal characteristics. Subsequently, we fine-tuned the pretrained model on HOT2020 and IMEC25 to adapt to their respective spectral properties and scene-specific differences. During fine-tuning, only the learning rate was reduced to one-tenth of the original value, while all other hyperparameters were kept unchanged.

	\noindent\textbf{Template Update:}
	The template update strategy can be expressed as:
	\begin{equation}
		\theta_t = \eta \theta_{t-1} + (1-\eta) s_t
	\end{equation}
	where $\theta_t$ is the accumulation threshold computed progressively over frames $1,2,\cdots, t-1$ and $\theta_t$ is initialized as $0$\textcolor{green}{.}  The paramete $\eta=0.7$ is a momentum factor to control the influence of the previous threshold. Here, $s_t$ is the tracking score of the current $t$-th frame, measured by the classification loss. The template is updated only when $s_t$ surpasses the current threshold $\theta_t$,  ensuring that updates are made with high certainty rather than low-scoring modifications. This adaptive strategy helps maintain a robust and accurate template, even in cases of occlusion or fast motion, by preventing premature or unreliable template updates.

\noindent\textbf{Evaluation Index:} All trackers are evaluated using the AUC of the precision plot and the distance precision (DP) at 20 pixels.  A precision plot shows the percentage of frames where the predicted bounding box is within a certain distance from the ground-truth bounding box. A success plot evaluates the overlap between the predicted and ground-truth bounding boxes, based on the Intersection over Union (IoU) metric.

\noindent\textbf{Compared Methods} We selected 12 methods to validate the effectiveness of our SUIT method, including BAE-Net~\cite{Li20BAE-Net}, MHT~\cite{Xiong20MHT}, SSDTNet~\cite{Lei2022SSDTNet},  TSCFW~\cite{Hou22TSCEW}, SEE-Net~\cite{Li23SEE-Net}, SiamBAG~\cite{li2023siambag}, SiamOHOT~\cite{Sun2023}, SPIRIT~\cite{Chen2024}, Trans-DAT~\cite{Wu2024},  SiamCAT~\cite{jiang2024siamcat}, PHTrack~\cite{Chen2024a}, and TBR-Net~\cite{Wang2024}. Except for MHT and TSCFW, which are based on hand-crafted features and run with CPUs, all other methods are deep learning-based and run with GPUs. By default, we directly used the given values in the original paper. Not all compared methods provide publicly available code, such as TBR-Net and SiamCAT, which are only tested on specific datasets and lack complete code for broader evaluation. This limitation prevents consistent testing across all datasets. Therefore, we conducted comparisons with available methods across different datasets where applicable.
\begin{figure}[tbp!]
	\centering
	\subfigure[Precision plot]{\includegraphics[width=0.48\linewidth, clip=true]{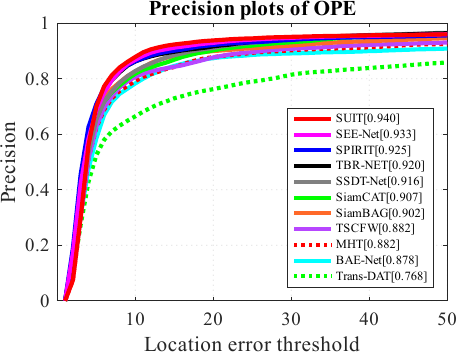}}
	\subfigure[Success plot]{\includegraphics[width=0.48\linewidth, clip=true]{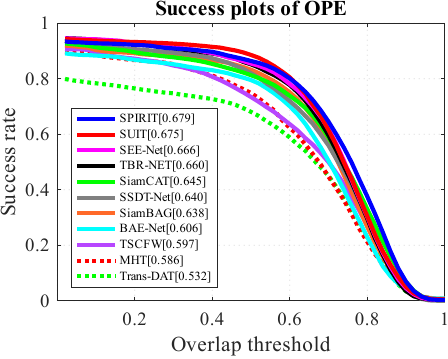}}
	\caption{Precision and success plots of all competing hyperspectral trackers on the HOT2020 dataset.} \label{fig:hot2020_auc_dp}
\end{figure}

\begin{figure}[ht]
	\begin{center}
		\includegraphics[width=\columnwidth]{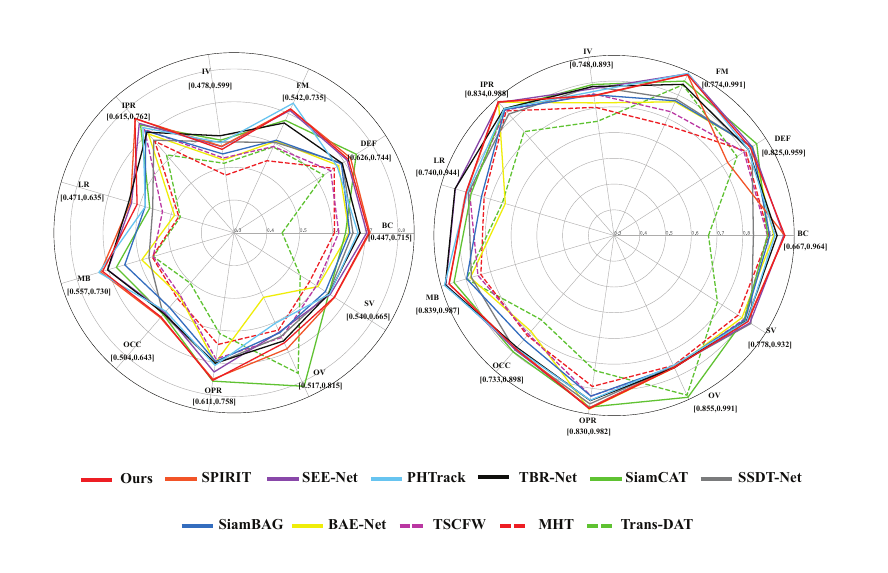}
	\end{center}
	\caption{Attribute-based comparison on the HOT2020 dataset.}
	\label{HOT20_Attribute}
\end{figure}
\begin{figure*}[ht]
	\begin{center}
		\includegraphics[width=\textwidth]{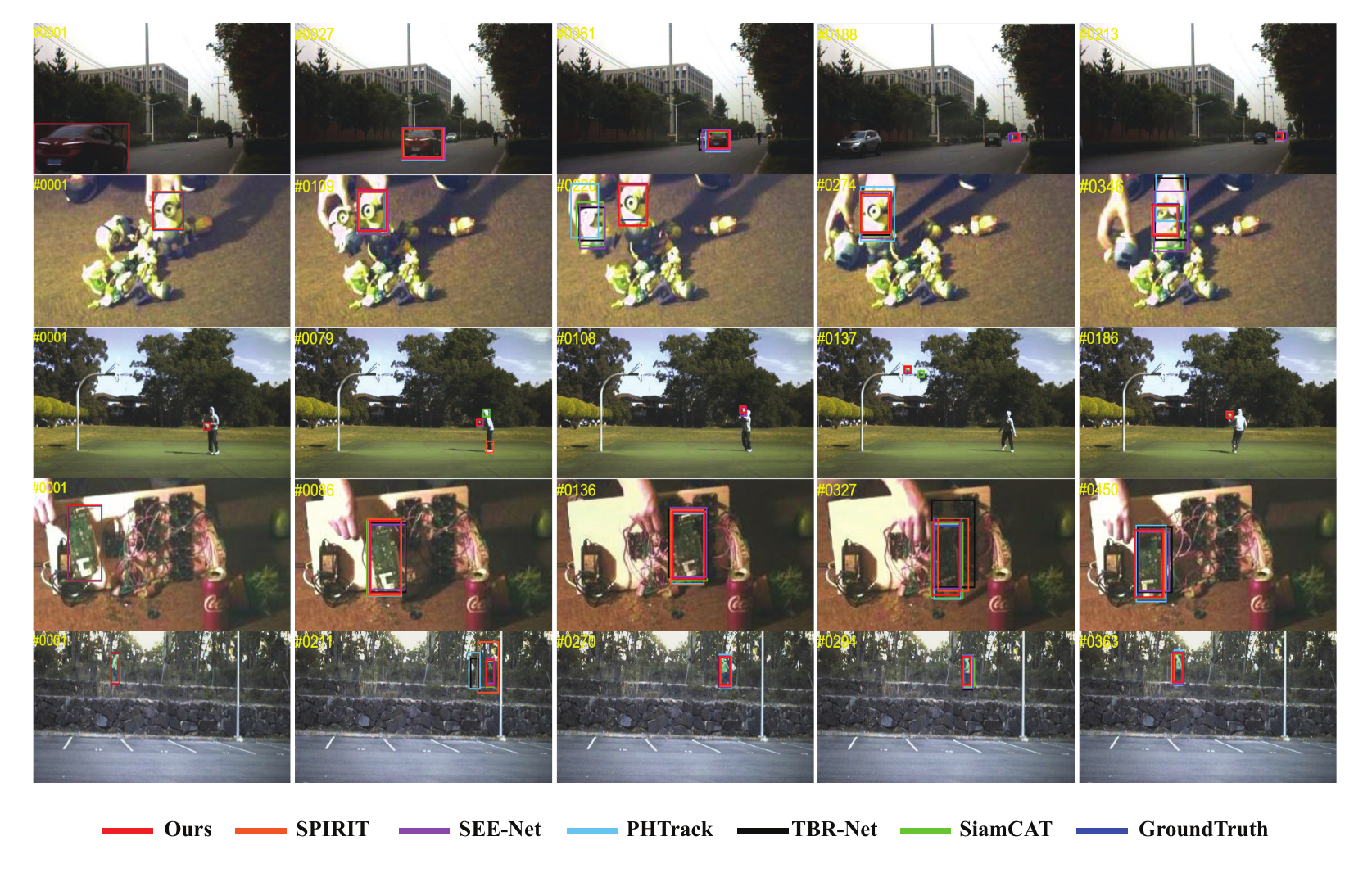}
	\end{center}
	\caption{Qualitative comparison of the SUIT on five video sequences (\emph{car3}, \emph{toy1}, \emph{basketball}, \emph{board}, and \emph{forest2}) on the HOT2020 dataset. The proposed approach demonstrates favorable performance against state-of-the-art trackers. For better visualization, we present the bounding boxes on false-color images converted from HSIs.}
	\label{HOT20_Visual}
\end{figure*}
	\subsection{Experiments with the HOT2020 Dataset}
	The HOT2020 dataset, provided by the Hyperspectral Object Tracking Competition (HOTC) in 2020~\footnote{Link: \url{https://www.hsitracking.com/}}, contains 16 bands in the visible range  from 460nm to 600nm. It includes 40 groups of videos for training and 35 groups of videos for testing, with an average of 390 frames per video. Each group features hyperspectral videos and false-color videos of the same scene,  converted from the hyperspectral videos. The HOT2020 dataset was annotated by 11 attributes corresponding to different challenging factors including Fast Motion (FM), Background Clutter(BC), Illumination Variation (IV), Out-of-Plane Rotation(OPR), Occlusion (OCC), Motion Blur (MB), Low Resolution (LR), Scale Variation (SV), In-Plane Rotation (IPR), Out-of-View (OV) and Deformation(DEF).

	Fig.~\ref{fig:hot2020_auc_dp} compares all the trackers on the HOT2020 dataset with respect to precision plots and success plots. Due to the inherent limitations of hand-crafted features in representing discriminative information, MHT and TSCFW perform worse than the other trackers. Thanks to advanced band relationship modeling with Transformer, TBR-Net is more effective at dividing hyperspectral frames into false-color frame groups for representation, resulting in improved tracking performance compared to SEE-Net, BAE-Net, and SiamBAG. Our SUIT method ranks top among all the approaches, achieving an AUC of \textbf{0.676} and a DP of \textbf{0.940}. The superior performance of SUIT can be attributed to two key factors. First, the effective spatial-spectral interaction enabled by the spatial-spectral union-intersection network and multihead attentions fully exploits both band-shared and band-specific information in HSVs for enhanced tracking. Second, the spectral loss function capitalizes on material consistency between the template and predicted regions, resulting in a more robust model capable of handling unexpected spatial structure changes. We also find that our SUIT slightly falls behind in the AUC index than SPIRIT. SPIRIT utilizes the initial template, search region, and dynamic template as inputs for spatial-spectral-temporal tracking. In contrast, our proposed SUIT only employs the search region and dynamic template. As a result, SUIT shows a slight disadvantage in terms of the AUC index compared to SPIRIT (0.676 vs. 0.679). However, benefiting from the spectral loss that enhances object localization by leveraging robust material information, SUIT achieves a notably higher performance in the distance precision index (0.940 vs. 0.925).
	

In Fig.~\ref{HOT20_Attribute}, we present a comparative analysis of various tracking methods based on several key properties. Our SUIT consistently ranks within the top three across all evaluated properties.  This exceptional performance can be attributed to the effective utilization of the spatial-spectral-temporal structure inherent in HSVs. Additionally, the spectral loss function enhances tracking, particularly in cases where spatial information is less reliable, as evidenced by the superior performance in OPR, IPR, and FM scenarios.

	Fig.~\ref{HOT20_Visual} presents a visual comparison of our SUIT tracker with competing methods on five representative sequences. Overall, SUIT demonstrates more precise predictions than other trackers, particularly in complex scenes where robust spatial-spectral interaction is crucial. For example, in the~\textit{basketball} sequence where the object is small and moves fast, most trackers, including TBR-Net and SPIRIT, lose the target, while the advanced spatial-spectral interaction in SUIT maintains more accurate tracking. In the \textit{forest2} sequence, the object shares similar color with the background.  SPIRIT and TBR-Net show significant deviations in their predictions, whereas SUIT remains stable tracking. Moreover, the spectral loss function in SUIT, which ensures material consistency between the predicted region and the template, enhances its robustness in sequences like~\emph{board} and~\emph{toy1}, where challenges such as rotation and background clutter exist.

\begin{figure}[t]
	\centering
	\subfigure[Precision plot]{\includegraphics[width=0.48\linewidth, clip=true]{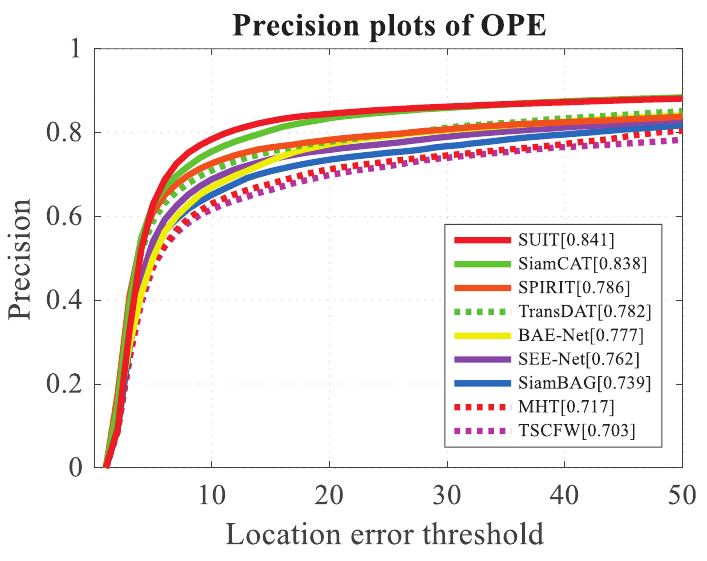}}
	\subfigure[Success plot]{\includegraphics[width=0.48\linewidth, clip=true]{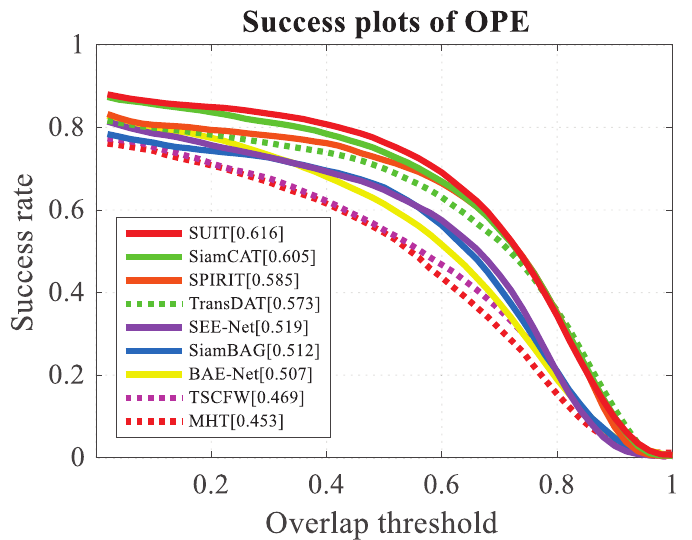}}
	\caption{Precision and success plots of all competing hyperspectral trackers on the HOT2023 dataset.} \label{fig:hot2023_auc_dp}
\end{figure}

\begin{figure}[!ht]
	\begin{center}
		\includegraphics[width=\columnwidth]{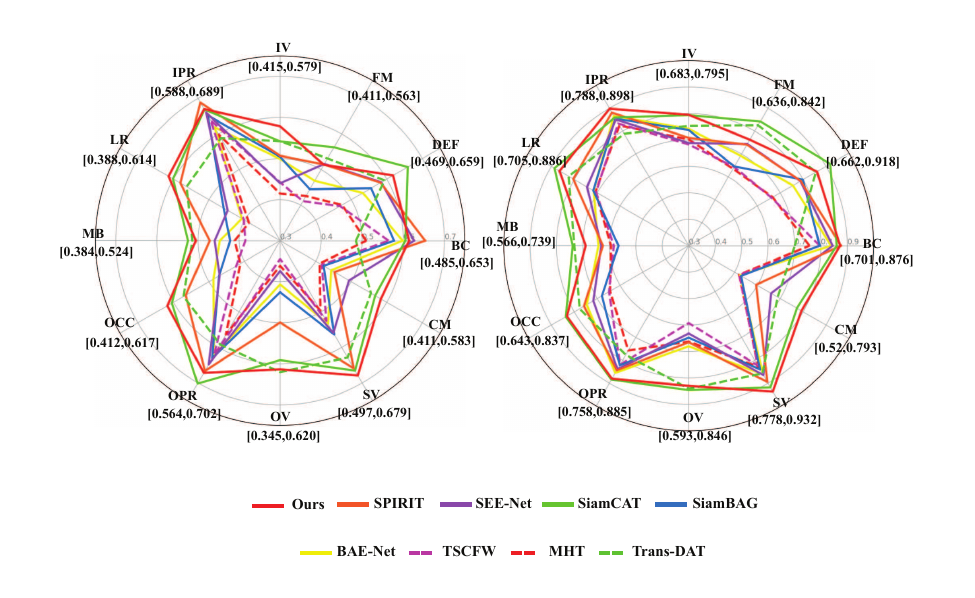}
	\end{center}
	\caption{Attribute-based comparison on the HOT2023 dataset.}
	\label{HOT23_Attribute}
\end{figure}

\begin{figure*}[ht]
	\begin{center}
		\includegraphics[width=\textwidth]{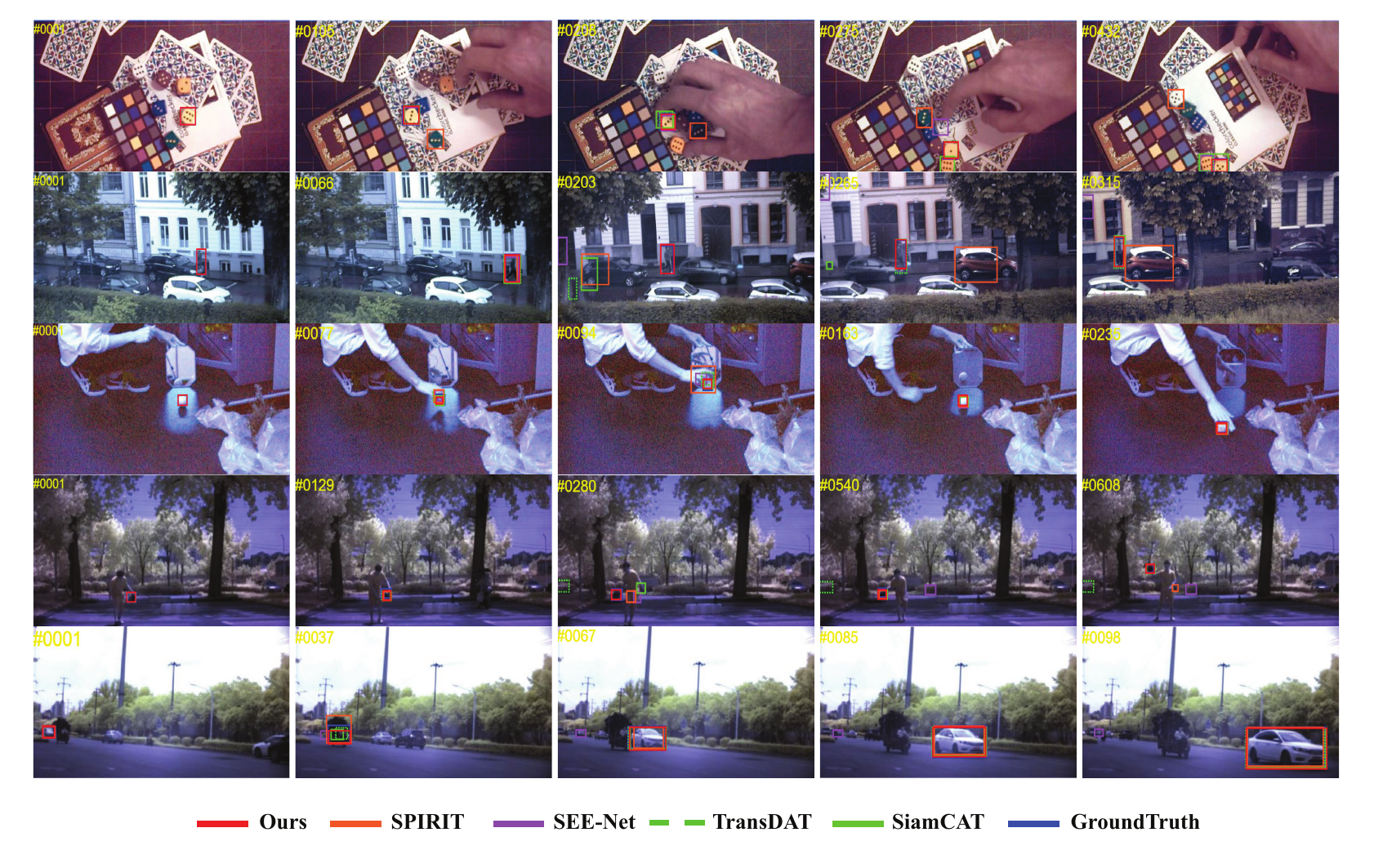}
	\end{center}
	\caption{Qualitative comparison of the SUIT on five video sequences (\emph{VIS-dice2}, \emph{VIS-rainystreet16}, \emph{RedNIR-ball-mirror9}, \emph{NIR-basketball3}, and \emph{NIR-car51}) on the HOT2023 dataset. The proposed approach demonstrates favorable performance against state-of-the-art trackers. For better visualization, we present the bounding boxes on false-color images converted from HSIs.}
	\label{HOT23_Visual}
\end{figure*}

		\subsection{Experiments with  HOT2023 Dataset}

		HOT2023 is a public dataset designed for evaluating hyperspectral object trackers in the third Hyperspectral Object Tracking Challenge in 2023. The dataset comprises a total of 108,000 frames across 161 videos in the visible range (16 bands from 460nm to 600nm), 100 videos in the near-infrared range (25 bands from 665nm to 960nm), and 41 videos in the red near-infrared range (15 bands from 600nm to 850nm). The dataset is further divided into 95 training videos and 77 testing videos. The varying number of bands and the unbalanced distribution across different modalities pose a significant challenge for developing a robust tracker. We excluded SiamOHOT and TBR-Net from the comparison due to unavailable tracking results and code.  Instead, we included  Trans-DAT, and SiamCAT into comparision. Except for SiamCAT, and our SUIT, which train a single model for all modalities, other deep learning-based approaches train three separate models, each corresponding to one type of video.
		
		In Fig.~\ref{fig:hot2023_auc_dp}, we present the performance of all the methods. Our SUIT outperforms the other compared methods by a significant margin, achieving the highest AUC of \textbf{0.615} and DP of \textbf{0.841}. Training multiple models for tracking is relatively easier, which is why SPIRIT and Trans-DAT achieve better performance in specific modalities. However, even with a single model for all modalities, our SUIT still delivers highly competitive performance, ranking in the top two across all methods for specific modalities. This success is primarily attributed to the introduced spatial-spectral union-intersection network and the spectral loss, which fully exploit the discriminative capabilities of HSVs for tracking.

		In addition to the 11 attributes included in HOT2020, the HOT2023 dataset also incorporates the Camera Motion(CM) attribute. We provide an analysis of tracking performance across 12 challenging attributes in Fig.~\ref{HOT23_Attribute}. Both SUIT and SiamCAT exhibit leading performance across various indicators, which may be attributed to their strategy of training a single model across three datasets, thereby facilitating better representation learning. Notably, SUIT surpasses SiamCAT in most cases concerning both the AUC and DP@20 indexes. This advantage stems from  SUIT's focus on comprehensive spatial-spectral interaction and maintaining material consistency.
		
		To further illustrate the advantages of the proposed SUIT, we present visual results from all trackers in Fig.~\ref{HOT23_Visual}. SUIT demonstrates superior performance in handling challenging scenarios such as SV~(\emph{NIR-car51}), BC~(\emph{VIS-dice2}), CM~(\emph{VIS-rainystreet16}), and LR~(\emph{VIS-dice2},~\emph{RedNIR-ball-mirror9}). In the~\emph{VIS-rainystreet16} sequence, challenges like out-of-view conditions and camera motion add complexity to the tracking task. Our SUIT addresses these issues by enforcing material consistency between the predicted and template regions, resulting in robust tracking. In LR scenarios, where capturing reliable spatial information is often difficult, effective spectral interaction and material consistency become crucial for enhancing tracking performance. For example, in the~\emph{RedNIR-ball-mirror9} sequence, our tracker maintains stable tracking, while methods like SPIRIT and SEE-Net exhibit varying degrees of tracking drift. In the~\emph{VIS-dice2} sequence, several frames show that all compared methods fail to track the target, yet our method consistently maintains accurate tracking.
		
		We have also recorded the runtime, number of parameters, and computational complexity of the compared methods on the HOT2023 dataset, as shown in Table~\ref{tab:tracker_comparison}. Specifically, we report these metrics for SUIT, SPIRIT, SEE-Net, and Trans-DAT, considering their representative performance and the availability of code. All experiments were conducted on a Linux machine equipped with an Intel Xeon CPU @ 2.40GHz, 128 GB RAM, and an NVIDIA RTX 3090 GPU.

		\begin{table}[h]
	\centering
	\caption{Comparison with respect to computational complexity, running time and number of parameters.} 
	\label{tab:tracker_comparison}
	\begin{tabular}{lccccccccccc}
		\toprule
		Tracker & SUIT &  SPIRIT & SEE-Net & Trans-DAT  \\
		\midrule
	    FPS & 8.75 &  14.32 & 8.7 & 33.2  \\
		FLOPS(G) & 258.4 & 52.43   & 244.2 & 30.28  \\
		Params(M) & 43.87 &  46.53 & 53.95 & 27.98\\
		\bottomrule
	\end{tabular}
\end{table}

Our SUIT model achieves a runtime of 8.75 frames per second (FPS), which is relatively slower compared to SPIRIT. This is primarily due to the additional computations introduced by the spatial-spectral interaction modeling. However, we believe that the trade-off in speed is justified by the significant improvements in tracking performance. Furthermore, with the increasing availability of cloud computing and high-performance hardware, the runtime of SUIT can be further optimized and reduced in practical deployments. Moreover, our SUIT has fewer parameters and less computational complexity while achieving higher performance, thanks to its more effective spatial-spectral correlation modeling, which helps unlock the full potential of HSVs.

		\subsection{Experiments with  IMEC25 Dataset}
\begin{figure}[t]
			\centering
			\subfigure[Precision plot]{\includegraphics[width=0.48\linewidth, clip=true]{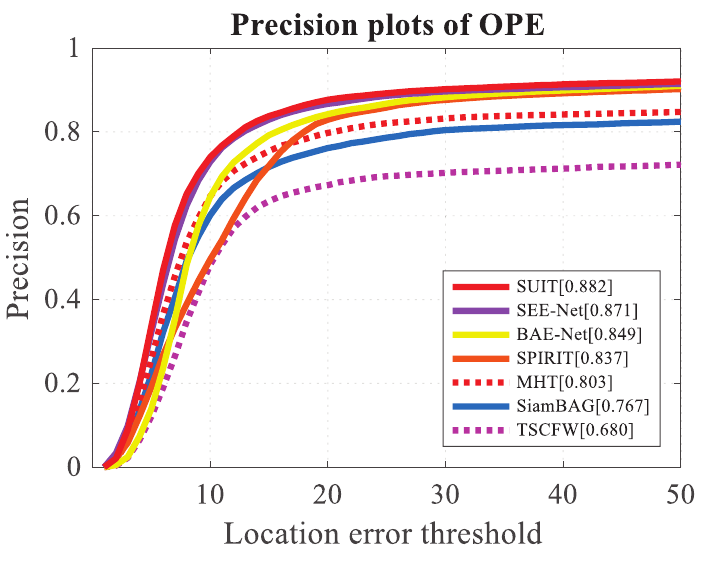}}
			\subfigure[Success plot]{\includegraphics[width=0.48\linewidth, clip=true]{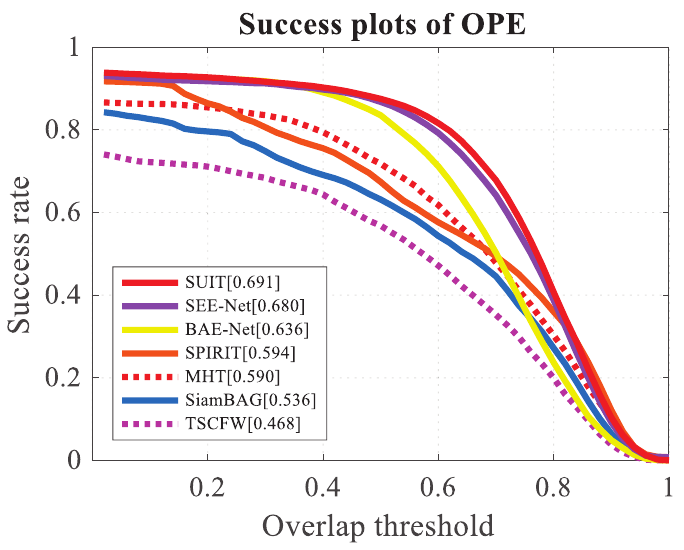}}
			\caption{Precision and success plots of all competing hyperspectral trackers on the IMEC25 dataset.} 	\label{fig:imec25_auc_dp}
		\end{figure}

		The IMEC25 dataset~\cite{Chen2022} includes 55 hyperspectral videos (HSVs) for training and 80 HSVs for testing. The training set comprises approximately 14,000 annotated frames, while the testing set contains around 14,060 annotated frames. Each frame in these HSVs includes 25 wavelength bands from 680 nm to 960 nm. As shown in Fig.~\ref{fig:imec25_auc_dp}, our SUIT still achieves the very competitive performance, showing its powerful tracking ability.
		
		\begin{figure}[!ht]
			\begin{center}
				\includegraphics[width=\columnwidth]{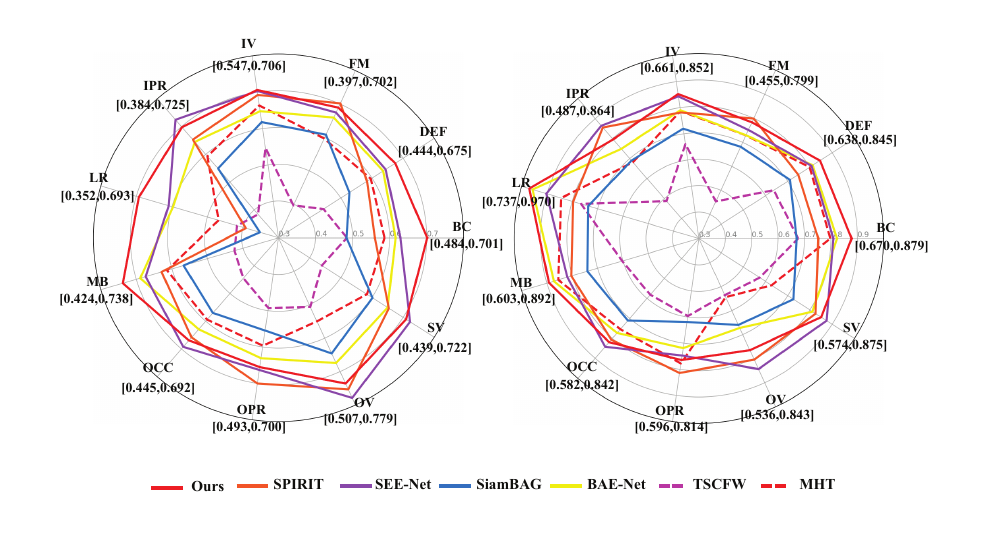}
			\end{center}
			\caption{Attribute-based comparison on the IMEC25 dataset.}
			\label{IMEC25_Attribute}
		\end{figure}

	\begin{figure*}[!ht]
			\begin{center}
				\includegraphics[width=\textwidth]{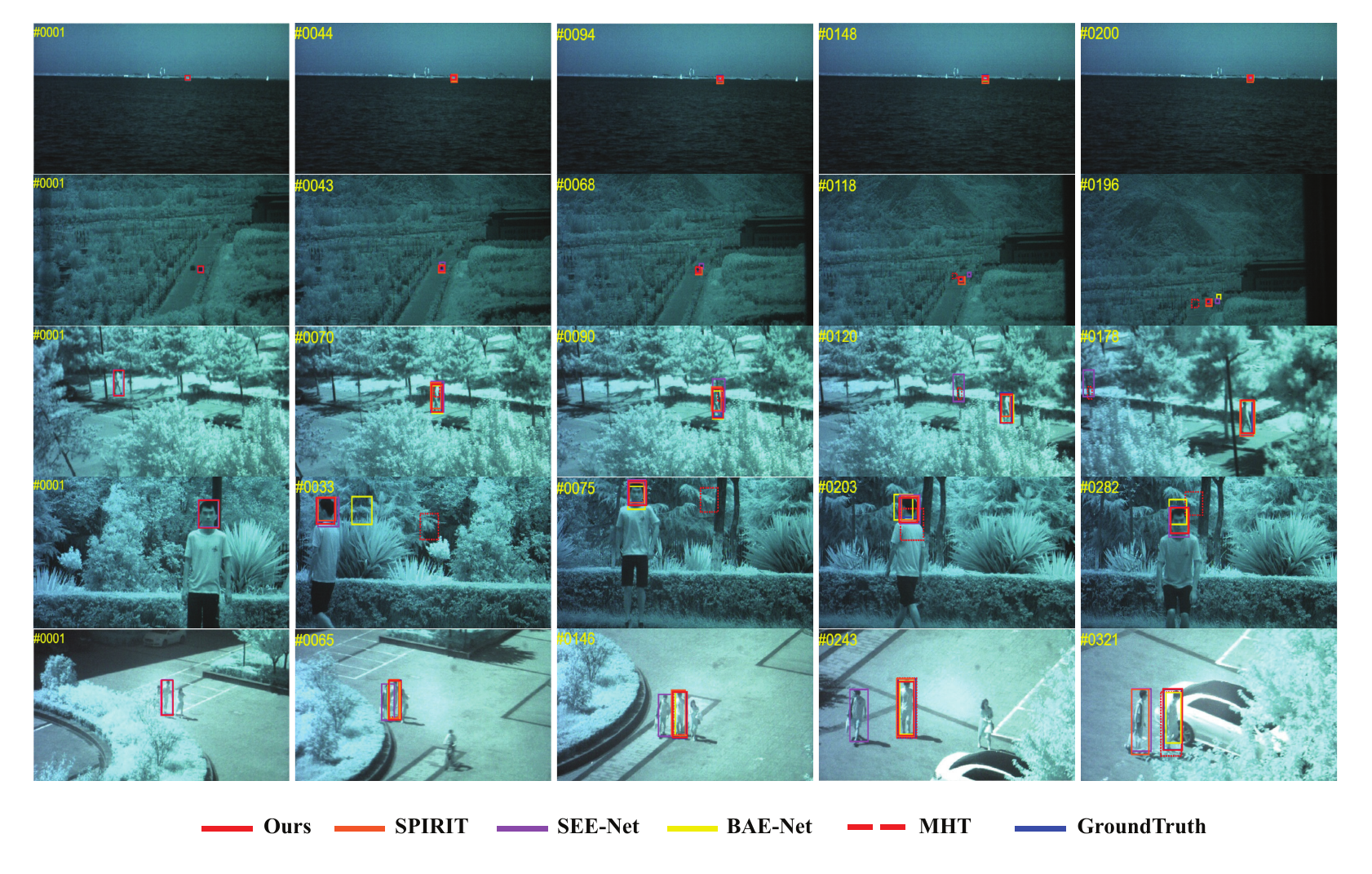}
			\end{center}
			\caption{Qualitative comparison of the SUIT on five video sequences (\emph{boat2}, \emph{doublecar}, \emph{human}, \emph{man3}, and \emph{triple}) on the IMEC25 dataset. The proposed approach demonstrates favorable performance against state-of-the-art trackers. For better visualization, we present the bounding boxes on false-color images converted from HSIs.}
			\label{IMEC25_Visual}
		\end{figure*}

		The IMEC25 dataset is annotated with the same 11 attributes as the HOT2020 dataset. As illustrated in Fig.~\ref{IMEC25_Attribute}, our SUIT consistently outperforms the compared trackers across various attributes. In the BC scenario, SUIT delivers a significant gain in the AUC index over SEE-Net by effectively integrating spatial and spectral information. Additionally, SUIT demonstrates notable advantages in the DEF, IV, and MB challenges. This superior performance further underscores the robustness and effectiveness of our SUIT approach.

		Since the IMEC25 dataset does not provide false-color videos, we generated false-color images by selecting the last three bands from the HSIs. In Fig.~\ref{IMEC25_Visual}, we present a visual analysis of five sequences: \emph{boat2}, \emph{doublecar}, \emph{human}, \emph{man3}, and \emph{triple}. When dealing with challenges like small targets with low resolution, such as in the \emph{boat2} and \emph{doublecar} sequences, most trackers lose the target. Our SUIT, however, integrates shared and specific information to construct an effective spatial-spectral interaction, resulting in more robust and stable tracking performance. Additionally, in scenarios with background interference from similar targets, as seen in the~\emph{triple} sequence, trackers like SEE-Net and SiamBAG fail, while our SUIT maintains superior robustness due to its consideration of spectral interaction. This experiment further demonstrates the effectiveness of the SUIT method in hyperspectral tracking.
		

		\subsection{Ablation Study}
		We present an ablation study on the spatial-spectral union-intersection network and the spectral loss in Table~\ref{Ablation}. All experiments were conducted on the HOT2023 dataset, given its large scale and diverse challenges.

\begin{table}[ht]
    \centering
    \caption{Effectiveness of the proposed key components.}
    \label{Ablation}
    \begin{tabular}{c c c c c c c}
        \toprule
        \textbf{Case} & \textbf{Sum} & \textbf{$f_{\text{shared}}$} & \textbf{$f_{\text{specific}}$} & \textbf{$\mathcal{L}_{\text{spec}}$} & \textbf{AUC} & \textbf{DP@20} \\
        \midrule
        1 & \cmark & \xmark & \xmark & \xmark & 0.535 & 0.701 \\
        2 & \xmark & \cmark & \xmark & \xmark & 0.586 & 0.806 \\
        3 & \xmark & \xmark & \cmark & \xmark & 0.599 & 0.815 \\
        4 & \xmark & \cmark & \cmark & \xmark & 0.610 & 0.821 \\
        5 & \xmark & \cmark & \cmark & \cmark & 0.615 & 0.841 \\
        \bottomrule
    \end{tabular}
\end{table}
	
		\begin{table}[ht]
			\centering
			\caption{Effectiveness of the parameter $\lambda_{spec}$ and $\eta$}
			\label{abla_param}
			\begin{tabular}{c>{\centering}p{3em}>{\centering}p{3em}>{\centering}p{3em}c}
				\toprule[1.2pt]
				\renewcommand{\arraystretch}{1.2}
				\setlength{\tabcolsep}{0.05em}
				\textbf{$\lambda_{spec}$}& n/a & 0.5 & 1 & 1.5  \\[3pt]
				\hline  \\[-1.5ex]
				\textbf{AUC}&0.610 & 0.612 & \textbf{0.615} & 0.613  \\
				\textbf{DP@20}&0.821 & 0.829 & \textbf{0.841} & 0.833  \\
				\toprule[1.2pt]
				\textbf{$\eta$}& 0.6 & 0.7 & 0.8 & 0.9 \\[3pt]
				\hline \\[-1.5ex]
				\textbf{AUC}&0.608 &  \textbf{0.615} & 0.604 & 0.599  \\
				\textbf{DP@20}&0.816 & \textbf{0.841} &0.812 & 0.808  \\
				\toprule[1.2pt]
			\end{tabular}
		\end{table}
		
		\noindent\textbf{Effectiveness of the Spatial-spectral Interaction:} In Case 1, replacing our SUIT with feature summation leads to a noticeable performance drop compared to Case 4. This suggests that direct addition lacks the ability of leveraging all the information to build spectral interactions between bands. Case 2 employs multiple SCF blocks based on the multihead cross-attention  to establish interactions among bands, resulting in higher performance than Case 1. However, due to its inability to extract band-specific information, the performance still falls short of Case 4. Similarly, Case 3, which focuses solely on extracting band-specific information  for interaction, also improves tracking performance compared to Case 1. In Case 4, both AUC and DP are significantly improved by simultaneously considering both band-common and band-specific interactions. Overall, the above phenomenon suggests that both band-specific and band-shared information should be taken into consideration for spectral-wise interaction.

		\noindent\textbf{Effectiveness of the Spectral Loss:}
		Building on the above analysis, the use of spectral loss $\mathcal{L}_{spec}$ further enhances performance, with AUC and DP increasing by 0.5\% and 2\%, respectively. This indicates that spectral loss helps maintain consistency between the template patch and the predicted patch in terms of material distribution. This is particularly beneficial in challenging cases where the spatial structure is unreliable, such as during object deformation and rotation.

		To further validate this, we conducted attribute-specific experiments, with a particular focus on attributes associated with significant shape variations, including DEF, IPR, OPR, and SV. As summarized in Table~\ref{tab:tracker_def_comparison}, the proposed SUIT model trained with the spectral loss consistently outperforms its variant trained without it. This performance gain indicates that the spectral loss provides a meaningful contribution even under severe deformation scenarios. The results further suggest that the spectral loss enhances robustness by leveraging spectral consistency as a supervisory signal, rather than relying solely on precise geometric boundaries, making the model less sensitive to shape distortions.

\begin{table}[h]
	\centering
	\caption{Attribute-based Performance comparison  with and without spectral loss.}
	\label{tab:tracker_def_comparison}
	\begin{tabular}{lcccc}
		\toprule
		Trackers & \multicolumn{2}{c}{With Spectral Loss} & \multicolumn{2}{c}{Without Spectral Loss}  \\
		\cmidrule(lr){2-3} \cmidrule(lr){4-5}
		Attributes & AUC & DP@20 & AUC & DP@20 \\
		\midrule
		DEF  & \textbf{0.618}  & \textbf{0.860}  & 0.595 & 0.836 \\
		IPR  & \textbf{0.670}  & \textbf{0.898}  & 0.654 & 0.860 \\
		OPR  & \textbf{0.672}  & \textbf{0.881}  & 0.648 & 0.849 \\
		SV   & \textbf{0.679}  & \textbf{0.936}  & 0.638 & 0.887 \\
		\bottomrule
	\end{tabular}
\end{table}

\begin{figure}[!ht]
	\begin{center}
		\includegraphics[width=\columnwidth]{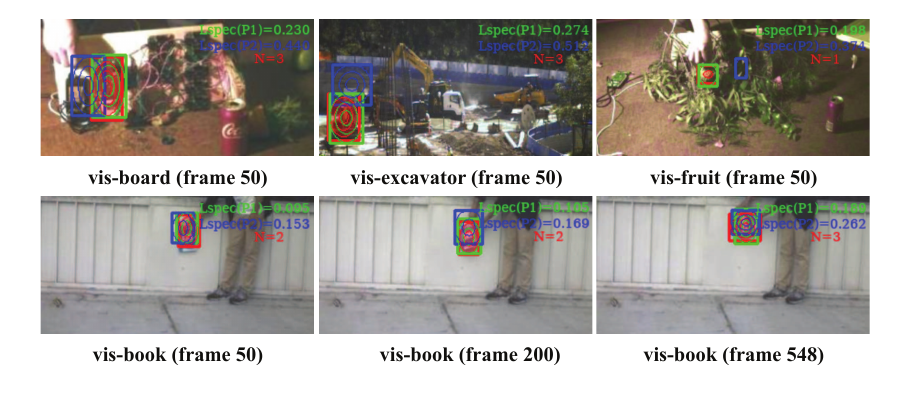}
	\end{center}
	\caption{Visualization of spectral loss behavior across different target shapes and deformations. Red bounding boxes denote the groundtruth regions; green and blue bounding boxes represent two candidate predicted regions.}\label{fig:spectral_loss}
\end{figure}

	Moreover, as illustrated in Fig.~\ref{fig:spectral_loss}, the spectral loss effectively adapts to targets with different shapes, sizes, and deformations. When the predicted bounding box overlaps more with the ground truth and captures regions with consistent material properties, the spectral loss significantly decreases, demonstrating its robustness to non-elliptical targets. The elliptical partitioning mechanism concentrates the focus on the core regions of the object rather than its entire shape, which is particularly beneficial for irregular objects such as~\emph{vis-fruit}. Furthermore, the number of ellipses $N$ is adaptively determined by the aspect ratio of the predicted box, enabling the loss to accommodate elongated or deformed objects, as shown in the \emph{vis-book} sequence. This adaptive design ensures that the spectral loss remains stable and meaningful even when the object undergoes shape variations or deformation.

\begin{figure}[!ht]
	\begin{center}
		\includegraphics[width=\columnwidth]{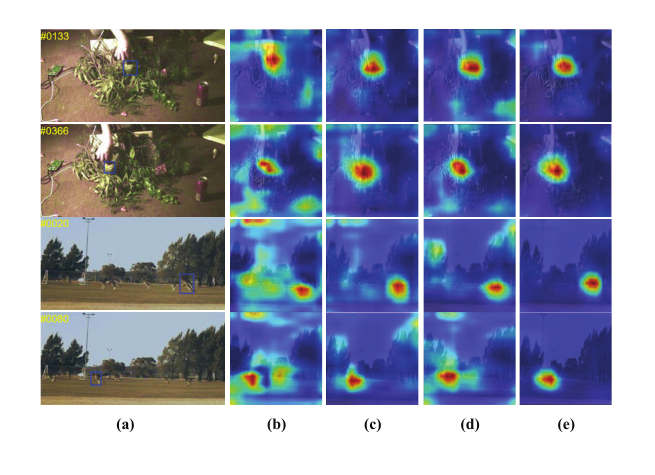}
	\end{center}
	\caption{Visualization of attention maps. (a) False-color images. (b) Attention maps from the spatial-wise interaction module without spectral loss. (c) Attention maps from the spectral-wise interaction module without spectral loss. (d) Attention maps from the spatial-wise interaction module with spectral loss. (e) Attention maps from the spectral-wise interaction module with spectral loss.}\label{fig:attention maps}
\end{figure}

	To gain further insight, we present the attention maps  under different settings with and without the spatial-spectral interaction module, and with and without spectral loss  in Fig.~\ref{fig:attention maps}. The comparison between (b, c) and (d, e) demonstrates that applying the inclusion–exclusion principle in the spectral-wise interaction enables the tracker to effectively suppress irrelevant background information and concentrate on the target region. Furthermore, comparing (b, d) and (c, e) illustrates that incorporating the spectral loss refines the training process, resulting in stronger target-focused attention.\\

		\noindent\textbf{Impact of  $\lambda_{spec}$ and $\eta$:}
		We present the impact of $\lambda_{spec}$ and $\eta$ on tracking performance in Table~\ref{abla_param}. For $\lambda_{spec}$, the DP@20 index is more sensitive to changes compared to the AUC as $\lambda_{spec}$ increases. This could be attributed to the spectral loss improving localization performance, particularly in scenarios involving object deformation and rotation, by enforcing material distribution consistency. However, achieving an accurate bounding box requires incorporating additional information, such as temporal cues. Optimal performance is observed at $\lambda_{spec}=1$.  The parameter $\eta$ controls the frequency of template updates. A larger $\eta$ results in frequent updates, while a smaller $\eta$ slows the update process. Optimal tracking performance is achieved at $\eta=0.7$, striking a balance between adapting to new appearances and maintaining stability.

\subsection{Failure Cases}
\begin{figure}[!ht]
	\begin{center}
		\includegraphics[width=\columnwidth]{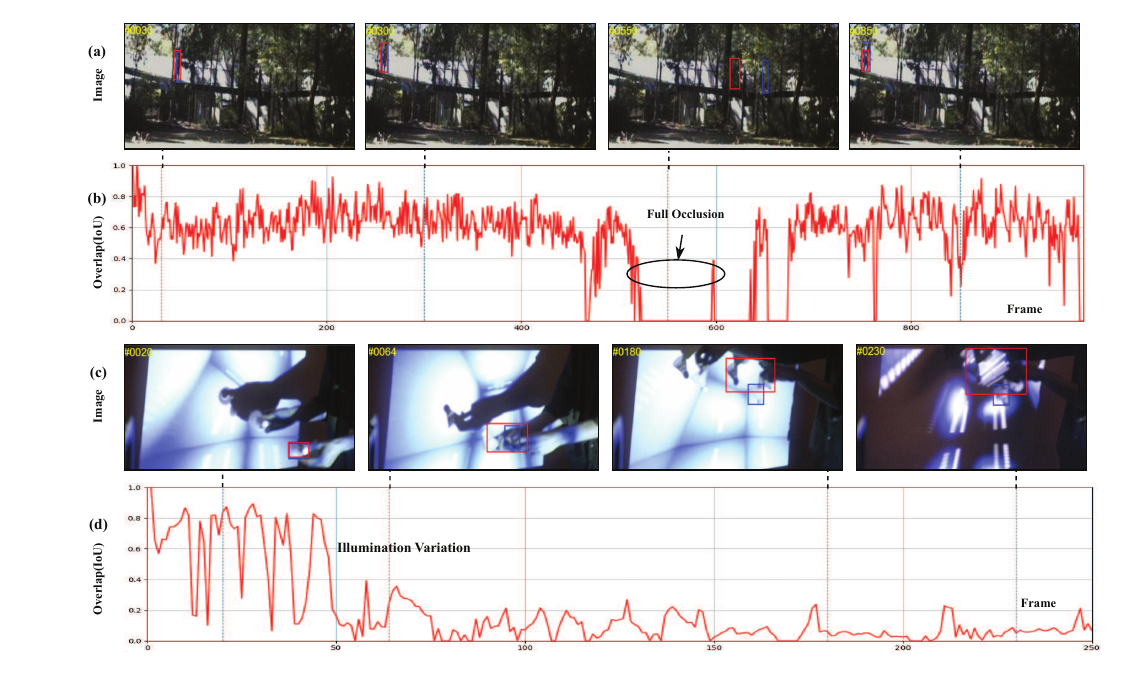}
	\end{center}
	\caption{Illustration of tracking failures on the  HOT2023 dataset. (a) and (c) shows  the \emph{VIS-campus} and \emph{VIS-partylight6} sequences, respectively, where the blue bounding box denotes the ground truth and the red bounding box indicates the prediction by SUIT. (b) and (d) show the corresponding overlap (IoU) curves.}\label{fig:failure_case}
\end{figure}

Here, we add representative failure cases in Fig.~\ref{fig:failure_case} to illustrate the limitations of our tracker in challenging scenarios. Specifically, in the case of long-term occlusion (e.g., around frame 600 in Fig.~\ref{fig:failure_case}(a)), the target becomes fully occluded for an extended period, resulting in temporary tracking failure and delayed recovery due to significant appearance changes. Similarly, in the sequence shown in Fig.~\ref{fig:failure_case}(c), severe illumination variations and strong shadows lead to early target loss (around frame 50), with limited success in re-localization. These examples underscore the current challenges in maintaining robustness under extreme occlusion and lighting conditions, emphasizing the need for future improvements in long-term memory mechanisms and illumination-invariant feature representations.

\section{Discussion} \label{sec:dis}

Despite the unique material identification capability of hyperspectral video  data, its full potential for robust object tracking remains under explored. In this paper, we investigate the spatial-spectral interaction between template and search regions and enforce material-distribution consistency. We show that fusing band-specific interactions through the inclusion-exclusion principle of set theory allows the tracker to leverage material cues more effectively, yielding a stronger spatial-spectral interaction. In addition, a material-guided loss further enhances robustness against unexpected shape changes. Our results confirm that spectral information is highly beneficial for object tracking.

However, our current design has two main limitations. First, it primarily focuses on spatial-spectral modeling and partially overlooks temporal correlations, reducing robustness in scenarios with long-term occlusions. Second, the computational cost limits real-time applicability. To address these issues, future work will: (i) incorporate temporal cues, possibly using the same inclusion-exclusion fusion strategy to improve resilience under occlusion, and (ii) develop dynamic band selection techniques to reduce complexity and enable real-time tracking.

		\section{Conclusions}\label{sec:con}
		
		This paper introduces a spatial-spectral interaction network for hyperspectral object tracking. By establishing both spatial-wise and spectral-wise relationships between the template and search regions, our network fully leverages the discriminative information in HSVs for robust tracking. Guided by the inclusion-exclusion principle, the spectral-wise interaction is more effectively constructed, significantly enhancing tracking performance. Additionally, the spectral loss further boosts performance by enforcing material consistency. It is noteworthy that our inclusion-exclusion principle-based multi-band interaction approach is quite general and can be applied for multiple modal fusion. In future work, we plan to extend this approach to multimodal tracking, such as RGBT and RGBD tracking, to further improve performance in these tasks.
		
		\appendices
		\bibliography{refs}
		\bibliographystyle{IEEEtran}
	\end{document}